\documentclass[twocolumn,
]{ceurart}

\usepackage{listings}
\usepackage{algorithm}
\usepackage{algpseudocode}
\usepackage{amsmath}
\usepackage{amsfonts}
\usepackage[title]{appendix}
\begin{document}

\copyrightyear{2025}
\copyrightclause{Copyright for this paper by its authors.
  Use permitted under Creative Commons License Attribution 4.0
  International (CC BY 4.0).}

\conference{RecSys in HR'25: The 5th Workshop on Recommender Systems for Human Resources, in conjunction with the 19th ACM Conference on Recommender Systems, September 22--26, 2025, Prague, Czech Republic.}

\title{Towards Explainable Job Title Matching: Leveraging Semantic Textual Relatedness and Knowledge Graphs}


\author{Vadim Zadykian}[
orcid=0009-0009-9749-6332,
email=vadim.zadykian@mymtu.ie,
]

\author{Bruno Andrade}[
orcid=0000-0002-1191-8082,
email=bruno.andrade@mtu.ie,
]

\author{Haithem Afli}[
orcid=0000-0002-7449-4707,
email=haithem.afli@mtu.ie,
]
\address{ADAPT Centre, Munster Technological University, Cork, Ireland}
        

\begin{abstract}
Semantic Textual Relatedness (STR) captures nuanced relationships between texts that extend beyond superficial lexical similarity. In this study, we investigate STR in the context of job title matching — a key challenge in resume recommendation systems, where overlapping terms are often limited or misleading. We introduce a self-supervised hybrid architecture that combines dense sentence embeddings with domain-specific Knowledge Graphs (KGs) to improve both semantic alignment and explainability. Unlike previous work that evaluated models on aggregate performance, our approach emphasizes data stratification by partitioning the STR score continuum into distinct regions: low, medium, and high semantic relatedness. This stratified evaluation enables a fine-grained analysis of model performance across semantically meaningful subspaces. We evaluate several embedding models, both with and without KG integration via graph neural networks. The results show that fine-tuned SBERT models augmented with KGs produce consistent improvements in the high-STR region, where the RMSE is reduced by 25\% over strong baselines. Our findings highlight not only the benefits of combining KGs with text embeddings, but also the importance of regional performance analysis in understanding model behavior. This granular approach reveals strengths and weaknesses hidden by global metrics, and supports more targeted model selection for use in Human Resources (HR) systems and applications where fairness, explainability, and contextual matching are essential.
\end{abstract}


\maketitle


\section{Introduction}
Semantic Textual Relatedness (STR) is a nuanced and context-dependent concept in Natural Language Processing (NLP) that measures the degree to which two text segments (words, sentences, or phrases) share semantically meaningful connections. Unlike Semantic Textual Similarity (STS) \cite{Chandrasekaran2021, Chen2024, Gaur2024}, which focuses primarily on surface-level closeness (e.g., synonyms or paraphrases), STR captures more abstract and associative relationships. For example, the words ``mitten'' and ``glove'' are semantically similar, whereas ``hand'' and ``glove'' are semantically related, yet dissimilar. STR also differs from Semantic Lexical Relatedness (SLR) \citep{Cruse1986, Webb2021}, which considers the relatedness of individual words rather than broader concepts.

While STS and SLR have been widely adopted in tasks such as paraphrase detection, question-answering, and summarization, STR remains underexplored, especially in domain-specific contexts such as talent acquisition and job recommender systems. In these settings, job titles often exhibit significant lexical diversity while denoting functionally similar or hierarchically related roles. Traditional keyword- or syntactic-based methods fail to account for this variation \cite{Zikanova2015, Webb2021}. For instance, ``Chief Executive Officer'' and ``Managing Director'' may have no shared tokens but represent nearly identical positions, whereas ``Director of Sales'' and ``Vice President, Marketing'' are distinct but related roles. This issue is compounded in global and multilingual hiring scenarios, where terminology is inconsistent or localized.

Another critical challenge in Human Resource (HR) applications is \textbf{explainability}. Job Recommender Systems (JRS) and Resume Recommender Systems (RRS) increasingly influence hiring decisions and career mobility. However, existing embedding-only methods such as Word2Vec \cite{Mikolov2013.Word2Vec} and BERT \cite{Devlin2019} often act as ``black boxes,'' producing relatedness scores without justifications. This lack of transparency undermines user trust and makes compliance with regulatory requirements problematic.

To address these challenges, we propose a hybrid framework that combines fine-tuned Sentence-BERT (SBERT) \cite{Reimers2019} embeddings with domain-specific Knowledge Graphs (KGs). This integration leverages the complementary strengths of embeddings and KGs: embeddings capture contextual semantics even in the absence of lexical overlap, while KGs encode structured hierarchical and functional relationships (e.g., career progressions, job categories). Critically, KGs enable us to trace explicit reasoning paths behind predictions (e.g., ``Project Lead'' $\rightarrow$ ``Team Leadership Roles'' $\rightarrow$ ``Program Manager''), thereby providing interpretability that is crucial in HR contexts.
Our work offers the following contributions:

\begin{itemize}
    \item we employ a self-supervised data pipeline that eliminates the need for manually labeled similarity scores by generating training pairs from cosine similarities between job descriptions
    \item we construct a knowledge graph to represent job-skill relationships and learn a neural mapping from textual embeddings to graph embeddings
    \item we introduce a hybrid modeling approach that integrates dense sentence embeddings with knowledge graph embeddings derived from a structured skill ontology    
    \item we utilize a stratified evaluation framework by partitioning similarity scores into three interpretable regions: low, medium, and high STR. This region-aware analysis enables fine-grained assessment of model behavior that would otherwise be obscured by global metrics such as RMSE or Pearson correlation
\end{itemize}

We hypothesize that STR-aware systems augmented with KGs will produce more diverse and contextually relevant job matches, while also meeting the explainability needs of HR stakeholders. This positions STR and KGs not only as technical improvements but also as key enablers of fairness, trust, and transparency in modern workforce ecosystems.


\section{Related Work}
\subsection{Semantic Textual Relatedness}
Semantic Textual Relatedness (STR) is often confused with Semantic Textual Similarity (STS) \cite{Chandrasekaran2021}. STS can be considered a component of STR \cite{Abdalla2021}, as semantically similar texts (e.g., paraphrases) are inherently related. However, STR encompasses a broader range of semantic associations beyond surface-level resemblance, including hierarchical, causal, and contextual connections \cite{Sloan1988,  Miyabe2008, Zikanova2015}.


\citet{Abdalla2021} demonstrate that integrating STR into search and recommendation pipelines enables retrieval of thematically relevant content, even when explicit term overlap is absent. 

Recommender systems utilizing STR generally outperform those that rely solely on surface similarity \citep{Lombardo2020}. These context-aware models can identify semantic connections between user preferences and items \citep{Ye2015}, improve personalization and diversity \citep{Likavec2015}, and mitigate cold-start problems \citep{Natarajan2022}. 

Recent advances in contextual language models have further propelled STR modeling. BERT \citep{Devlin2019} and Sentence-BERT (SBERT) \citep{Reimers2019} effectively capture polysemy and co-reference by leveraging bidirectional context. 

Within the context of JRS and RRS, various BERT-based models are used for job title classification \citep{Rahhal2023}, resume classification \citep{Tanberk2023}, job-resume matching \citep{Pias2024, Kaya2023, Rezaeipourfarsangi2023,Rosenberger2025}, Named Entity Recognition \citep{Girish2023, Dhobale2025, Tanberk2023}, and semantic ranking of job recommendations \citep{Tang2024, Ramyar2024, Aleisa2023}.

In summary, STR has emerged as a key factor in the development of intelligent systems that require deep semantic understanding. Contextual language models, especially fine-tuned SBERT variants, provide a solid foundation for approximating STR \citep{Abdalla2021, Zhao2020, Misra2020, Zou2022, Hammami2024}. 

\subsection{Explainability in HR Systems}
Transparency in HR systems is not only desirable, but it may also be mandatory. For example, the EU AI Act \cite{eu_ai_act_2025} classifies certain systems used in HR as “high-risk” and subjects them to strict traceability and explainability requirements. 
Explainability in HR systems, particularly in job recommendation systems, is indispensable for fostering fairness, transparency, loyalty and trust \cite{Min2022, 10.1561/1500000066, Bied2023}, facilitating informed decision-making \cite{10.1561/1500000066}, mitigating biases \cite{10.52783/jes.1721, 10.3390/app132212305}, and catering to diverse stakeholder needs \citep{10.1145/3640457.3688014, Bied2023, 10.1007/978-3-031-44067-0_30}, while ensuring regulatory compliance. 

\subsection{Knowledge Graphs}
Knowledge graphs (KGs) have emerged as valuable tools for enhancing the explainability of recommender systems. 

Recent works have demonstrated how combining KGs with pre-trained language models (PLMs) — such as BERT and SBERT — can improve both knowledge graph completion and downstream semantic tasks \citep{Yao2019.KGBERT, Liu2022.KGBERT, Xu2023.KGBERT}. 

Building on these insights, our approach leverages alignment between textual embeddings and structured knowledge graphs to improve job-to-job and job-to-skill similarity estimation. Inspired by multi-task frameworks \citet{Kim2020-KGBERT} and contextualized reasoning models \citep{Gul2024.KGBERT}, we aim to train sentence encoders that not only capture linguistic nuance but are also sensitive to the structural topology of skills and industries. 

\subsection{Job Title Matching}

Several recent studies have explored the use of representation learning and transformer-based models in the context of job matching and job title normalization. \citet{Zhang2019.Job2Vec} introduce Job2Vec, a multi-view framework that learns job title embeddings by integrating structured and unstructured job-related data. \citet{Lavi2021A} propose conSultantBERT, a fine-tuned Siamese SBERT model, which improves job–candidate matching over keyword-based approaches by leveraging domain-specific data. Building on similar ideas, \citet{Kaya2021.effectiveness} investigate sentence-pair classification models using job titles as input signals and highlight the effectiveness of fine-tuned BERT embeddings for resume-to-job matching\cite{Kaya2023}.

\citet{Decorte2021} explore job title normalization using BERT-based models fine-tuned on recruitment corpora, showing improved classification into standardized taxonomies. \citet{Liu2022.Title2vec} propose Title2Vec, a job title embedding approach designed for Named Entity Recognition (NER) and classification tasks. \citet{Zbib2023} introduce a weakly supervised method for learning job title similarity by mining noisy skill co-occurrence patterns, offering a scalable alternative to manually labeled training data. In a related effort, \citet{Rosenberger2025} present CareerBERT, a transformer-based architecture that aligns resumes and ESCO job descriptions for job recommendation tasks.

Collectively, these studies demonstrate the growing relevance of contextual and self-supervised embeddings in addressing real-world challenges in job matching, normalization, and recommendation. Our work builds upon these foundations and contributes further by integrating semantic representations with knowledge graph embeddings.


\section{EXPERIMENTAL FRAMEWORK}

\subsection{Motivation and Research Objectives}

Job Recommender Systems leverage a variety of input signals, including unstructured text (e.g., resumes or job postings), structured data (e.g., user preferences and skills), and collaborative features (e.g., interaction history between users and job listings). One commonly available yet often underutilized signal is the job title, which has been shown to carry meaningful semantic information \cite{Kaya2021.effectiveness}.

While relying solely on job titles for matching would be insufficient, understanding the semantic textual relatedness (STR) between job titles can enhance filtering, refinement, and personalization in recommendation systems. This motivates our exploration of how job titles relate to one another and how these relationships can be explained —- contributing to more transparent and informed recommendations \cite{10.1561/1500000066}.

\subsubsection{Research Objectives}
\begin{itemize}
    \item Develop a self-supervised approach for extracting semantic representations of skills and job functions.
    \item Automatically generate labeled datasets for training and evaluation.
    \item Assess text embedding strategies in capturing semantic relationships between job titles.
    \item Evaluate graph-based models for their ability to represent and explain job title relationships.
    \item Analyze model performance across the full STR spectrum.
\end{itemize}

\subsection{Proposed Method}
We propose a self-supervised pipeline for learning semantic representations of job titles and their alignment with skill knowledge graphs. The description of the pipeline is presented in Algorithm~\ref{alg:Pipeline}.

\begin{algorithm}
\caption{Self-Supervised Semantic Job Embedding and Skill Mapping Pipeline}\label{alg:Pipeline}
\begin{algorithmic}[1]
\State \textbf{Input:} Raw job titles \& descriptions, skill descriptions
\State \textbf{Output:} Trained job title to skill graph alignment model

\State \textbf{Step 0: Summarization} \\
\hspace{1em} Apply a pretrained BART model \cite{Lewis2020.BART} to summarize each job description, removing boilerplate and retaining functional content.

\State \textbf{Step 1: Job Embedding Generation} \\
\hspace{1em} Encode each summary using SBERT to obtain auxiliary job embeddings.

\State \textbf{Step 2: Pairwise Similarity Computation} \\
\hspace{1em} Compute cosine similarity between all job embeddings to generate relatedness scores.

\State \textbf{Step 3: STR Dataset Construction and SBERT Fine-Tuning} \\
\hspace{1em} Construct a self-supervised dataset using similarity scores. Split into train/eval with disjoint job titles. Fine-tune SBERT on the training set.

\State \textbf{Step 4: Skill Embedding Generation} \\
\hspace{1em} Encode textual descriptions of skills using a transformer model to obtain skill embeddings.

\State \textbf{Step 5: Extraction of Job Functions and Skills} \\
\hspace{1em} For each job, compute cosine similarity to skills. Select top-ranked skills as semantic matches.

\State \textbf{Step 6: Knowledge Graph Construction and Embedding} \\
\hspace{1em} Construct a bipartite graph of jobs and related skills. Learn node embeddings using a graph embedding model (e.g., RGCN, ComplEx).

\State \textbf{Step 7: Embedding Alignment} \\
\hspace{1em} Train a neural network to map SBERT job title embeddings to the graph embedding space.

\State \textbf{Return:} Fine-tuned SBERT model and trained graph model.

\end{algorithmic}
\end{algorithm}

Building upon insights from prior studies \citep{Kim2020-KGBERT, Lavi2021A, Kaya2021.effectiveness, Decorte2021, Liu2022.Title2vec, Zbib2023, Xu2023.KGBERT, Gul2024.KGBERT, Rosenberger2025}, our approach introduces several methodological innovations in the domain of job title similarity and normalization. We extend existing work by integrating self-supervised learning with pretrained language models and leveraging a skill-centric knowledge graph to enhance interpretability.

Departing from the Job2Vec framework \citep{Zhang2019.Job2Vec}, which relies on co-occurrence and structural signals, we adopt a self-supervised strategy that aligns contextual embeddings with knowledge graph embeddings built around explicit job-skill relationships. Unlike \citet{Lavi2021A} and \citet{Rosenberger2025}, whose models emphasize resume-job alignment or broad job matching, our focus is specifically on job title normalization and similarity estimation, leveraging both textual and structured graph-based representations.

While \citet{Decorte2021} pursue supervised classification into a predefined taxonomy, our methodology emphasizes self-supervised learning techniques that aim to capture fine-grained semantic and structural relationships between job titles and skills. Similarly, although we share with \citet{Kaya2021.effectiveness} the use of job titles as primary input signals, we extend this by incorporating richer semantic cues from full job descriptions. In comparison to \citet{Liu2022.Title2vec}, our model embeds a broader semantic scope by integrating additional contextual, relational, and graph-based information.

Finally, we align with the objective of \citet{Zbib2023} in leveraging weak supervision for job title similarity, but diverge in our implementation by combining pre-trained sentence encoders with a structured knowledge graph of job-skill relationships. This enables our model to move beyond raw skill co-occurrence signals and instead produce semantically aligned, interpretable embeddings that support robust similarity estimation across diverse job roles.

\subsection{Text Vectorization Models}
We evaluate five different text vectorization model configurations which are summarized in Table \ref{table:EXP_VECTORIZATION_PARAMETERS}.

\begin{table}[h!]
\centering
\begin{tabular}{|l|p{3.5cm}|c|}
\hline
\textbf{Vectorizer}  & \textbf{Model Description} & \textbf{Size} \\
\hline

JOBBERT &  Pre-trained JOBBERT model \cite{Decorte2021} based on the SBERT architecture {\textquoteleft}jjzha/jobbert-base-cased{\textquoteright} & 768\\
\hline
JOBBERT-F & JOBBERT model {\textquoteleft}jjzha/jobbert-base-cased{\textquoteright}, fine-tuned on synthetic data & 768\\
\hline

MPNET & Pre-trained MPNET model based on the SBERT architecture {\textquoteleft}all-mpnet-base-v2{\textquoteright}.  & 768\\
\hline

MPNET-F & MPNET {\textquoteleft}all-mpnet-base-v2{\textquoteright}, fine-tuned on synthetic data & 768\\
\hline

MPNET+RGCN & Fine-tuned MPNET model {\textquoteleft}all-mpnet-base-v2{\textquoteright}, coupled with a Graph R-GCN model \cite{Schlichtkrull2017.RGCN}. R-GCN was selected because it is designed for node-level tasks (e.g., classification, regression). & 500\\
\hline
\end{tabular}
\caption{Vectorization models evaluated for the STR learning task}
\label{table:EXP_VECTORIZATION_PARAMETERS}
\end{table}

We evaluate every model on the same validation dataset which contains diverse job title pairs not included in any of the training data. For evaluation of STR prediction, we adopt Root Mean Squared Error (RMSE). This choice is valid because we want to optimize for precise similarity scores, and not just relative order. 

\subsection{Implementation Details}
The experiments were carried out in the Google Colab environment \cite{Google.Colab} with specifications listed in Table~\ref{table:EXP_SYSTEM_SPEC} (see Appendix~\ref{AppendixA}). The proposed pipeline requires several key hyper-parameters which are specified in Table~\ref{table:EXP_HYPER_PARAMETERS} (see Appendix~\ref{AppendixA}). 
The implementation code was written in Python \citep{python} using Visual Studio \citep{visualstudio}. The source code, together with the input and output files, is available at \citep{jobtitle_relatedness_2025}.

\subsubsection{Mapping Text to KG Space}

\paragraph{Training.}
We fine-tune the SBERT model using anchor-sample-score triplets and cosine similarity loss. We learn a parametric mapping that projects a job title’s SBERT embedding into the knowledge-graph (KG) embedding space. We use a lightweight MLP with $\ell_2$ normalization and MSE embedding loss. 

\paragraph{Inference.}
At inference time, we encode job titles via fine-tuned SBERT to obtain text embedding vectors, then compute graph embedding vectors, and calculate cosine similarity to estimate the STR score.

\subsubsection{Skill Selection and Graph Pruning}

To prevent generic skills from dominating the graph and explanations, we remove skills with job share $>20\%$. Remaining skills are reweighted by a specificity score (inverse centrality degree). This yields a sparser, more discriminative KG and improves both retrieval quality and explanation specificity.

\subsection{Stratified Data Sampling}
\begin{figure}[h!]
\centering
\includegraphics[width=7cm]{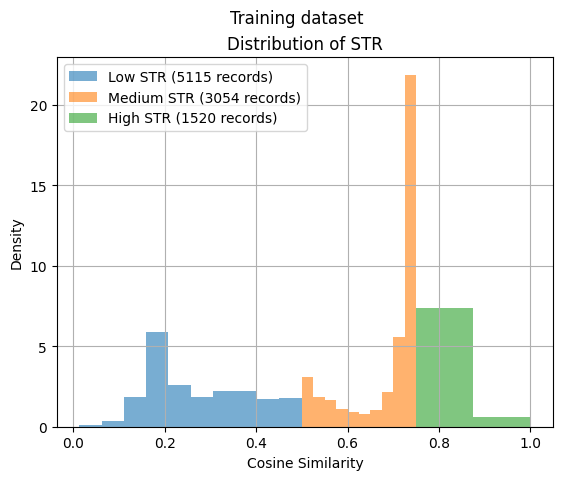}
\caption{Distribution of STR in Training dataset}
\label{fig:TRAIN_STR_DISTRIBUTION}
\end{figure}

To better understand and evaluate model behavior across varying levels of semantic textual relatedness (STR), we partition the continuous STR range into three semantically meaningful zones (Figure~\ref{fig:TRAIN_STR_DISTRIBUTION}), guided by domain expertise in HR:
\begin{itemize}
    \item Low STR (0.0–0.50): Pairs of job titles that are largely unrelated or noisy, often representing very different occupations or sectors.
    \item Medium STR (0.50–0.75): Ambiguous or borderline cases, which tend to be more challenging due to partial overlap in semantics.
    \item High STR (0.75–1.0): Highly related job title pairs, including near-duplicates, synonyms, or slight variations of the same role.
\end{itemize}

This stratification enables a more nuanced evaluation of model performance, as global performance metrics (e.g., overall RMSE or Pearson correlation) may obscure variation in model behavior across different semantic regions. We hypothesize that some models will perform better in specific STR regions while underperforming in others. For example, models trained with Cosine Similarity Loss may effectively distinguish highly similar and dissimilar titles but struggle with borderline cases.

\subsection{Data}
The raw data are obtained from several open-source collections: a Kaggle dataset \citep{Koneru2024}, a list of granular skills and competencies by ESCO \cite{EuropeanCommission2021ESCO}, and a list of broad job functions Indeed job site \cite{Indeed2025.JobTitles}. Table~\ref{table:DATA_FILES} (see Appendix~\ref{AppendixA}) describes the data files consumed and produced by the system.


\section{Results \& Discussion}

Each vectorization model is evaluated on the same validation dataset where the predicted STR is calculated using the cosine distance between two final embedding vectors. We calculate the global RMSE as well as RMSE values for every data region of interest (i.e. Low STR, Medium STR, High STR), which are presented in Table \ref{table:SIMILARITY_TASK_RESULTS}.

We also perform paired t-tests to reveal how each model's performance varies across STR regions (Low, Medium, High) based on absolute errors (Table~\ref{table:RESULT_TTESTS} and Figure~\ref{fig:RESULT_TTEST_HEATMAP}).

\begin{table}[h!]
\centering
\begin{tabular}{|l|c|c|c|c|}
\hline
 & \textbf{Global} & \textbf{Low} & \textbf{Medium} & \textbf{High} \\
\textbf{Vectorizer}  & \textbf{STR} & \textbf{STR} & \textbf{STR} & \textbf{STR} \\
\hline
JOBBERT	&	0.28	&	0.38	&	\textbf{0.11}	&	0.15	\\
\hline
JOBBERT-F &	0.17 & 0.16 & 0.17 &	0.18	\\
\hline
MPNET	&	0.29	&	\textbf{0.14}	&	0.36	&	0.44	\\
\hline
MPNET-F	&	\textbf{0.16}	&	\textbf{0.14}	&	0.16	&	0.18	\\
\hline
MPNET+RGCN	&	0.23	&	0.30	&	{0.14}	& \textbf{0.11}	\\
\hline
\end{tabular}
\caption{RMSE values across STR regions}
\label{table:SIMILARITY_TASK_RESULTS}
\end{table}

\begin{table}[h!]
\centering
\begin{tabular}{|l|l|r|r|l|}
\hline
\textbf{Model} & \textbf{Comparison} & \textbf{t-value} & \textbf{p-value}\\
\hline
JOBBERT     & Low \textit{vs} Medium   & 28.40  & 0.00 \\
            & Medium \textit{vs} High  & -4.61  & 0.00 \\
            & Low \textit{vs} High     & 23.04  & 0.00 \\
\hline
JOBBERT-F   & Low \textit{vs} Medium   & -2.11  & 0.04 \\
            & Medium \textit{vs} High  & -0.12  & 0.90 \\
            & Low \textit{vs} High     & -2.11  & 0.04 \\
\hline
MPNET       & Low \textit{vs} Medium   & -19.38 & 0.00 \\
            & Medium \textit{vs} High  & -5.57  & 0.00 \\
            & Low \textit{vs} High     & -24.21 & 0.00 \\
\hline
MPNET-F     & Low \textit{vs} Medium   & -2.23  & 0.03 \\
            & Medium \textit{vs} High  & -0.90  & 0.37 \\
            & Low \textit{vs} High     & -3.03  & 0.00 \\
\hline
MPNET-RGCN  & Low \textit{vs} Medium   & 20.43  & 0.00 \\
            & Medium \textit{vs} High  & 3.72   & 0.00 \\
            & Low \textit{vs} High     & 23.77  & 0.00 \\
\hline
\end{tabular}
\caption{Paired t-test results for absolute errors across STR regions}
\label{table:RESULT_TTESTS}
\end{table}


\subsection{Evaluation of Region-Specific Model Behavior}
\begin{figure}[h!]
\centering
\includegraphics[width=7cm]{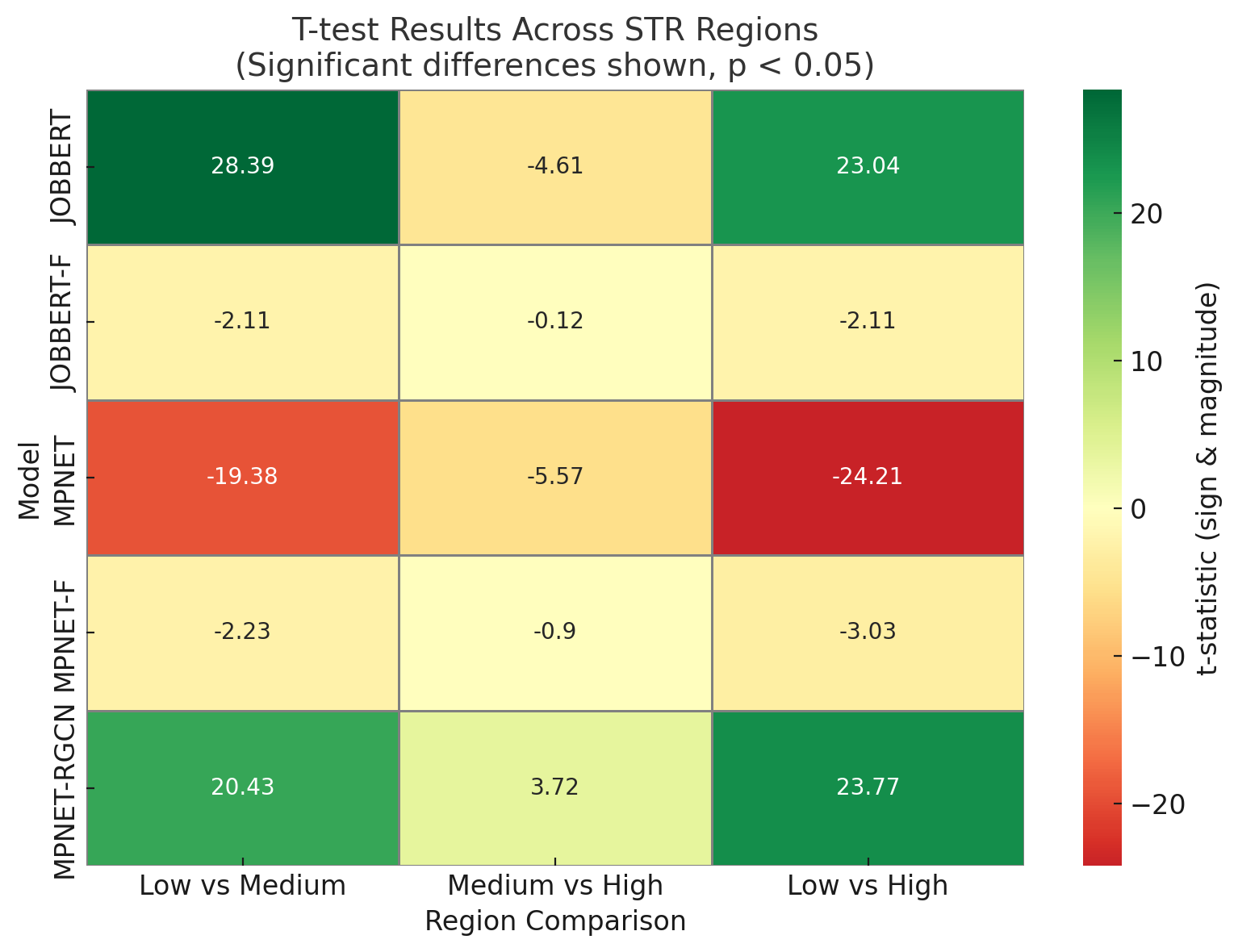}
\caption{T-Test Results Across STR Regions}
\label{fig:RESULT_TTEST_HEATMAP}
\end{figure}

Our analysis highlights the limitations of aggregate metrics such as global RMSE, which may mask meaningful variation in model performance across different STR zones, as demonstrated in Table~\ref{table:SIMILARITY_TASK_RESULTS}. Stratified evaluation reveals that models often exhibit asymmetric performance as evident from Figure~\ref{fig:RESULT_TTEST_HEATMAP}. A model may perform reliably in distinguishing unrelated job titles (low STR), yet be less consistent in matching similar roles (medium to high STR).

The paired t-test analysis reveals statistically significant differences in model performance across STR regions, as measured by absolute errors. 

\subsubsection{JOBBERT} 
\begin{figure}[!htbp]
\centering
\includegraphics[width=7cm]{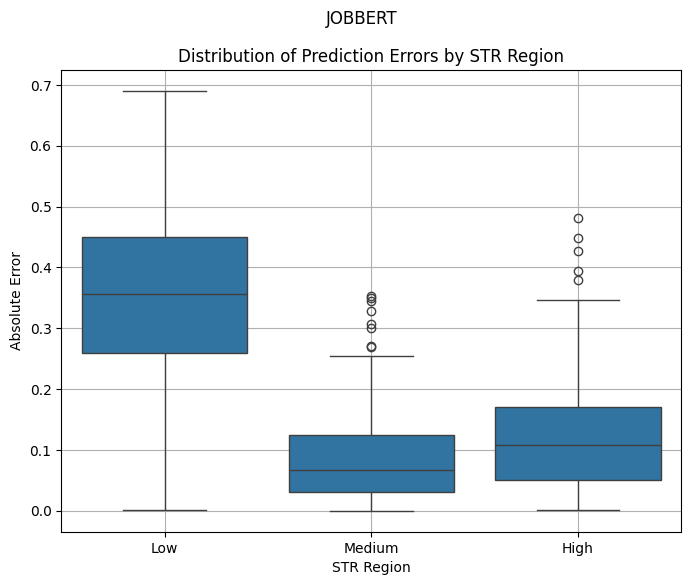}
\caption{Distribution of Prediction Errors for JOBBERT}
\label{fig:EVAL_ERROR_JOBBERT_BOXPLOT}
\end{figure}

JOBBERT demonstrates positive t-statistics in ``Low vs Medium'' and ``Low vs High'' which suggests that the model performs better in Medium and High STR bands compared to Low. Conversely, the negative t-statistic in ``Medium vs High'' implies superior performance in the Medium band (Figure~\ref{fig:EVAL_ERROR_JOBBERT_BOXPLOT}).

\subsubsection{JOBBERT-F}
\begin{figure}[!htbp]
\centering
\includegraphics[width=7cm]{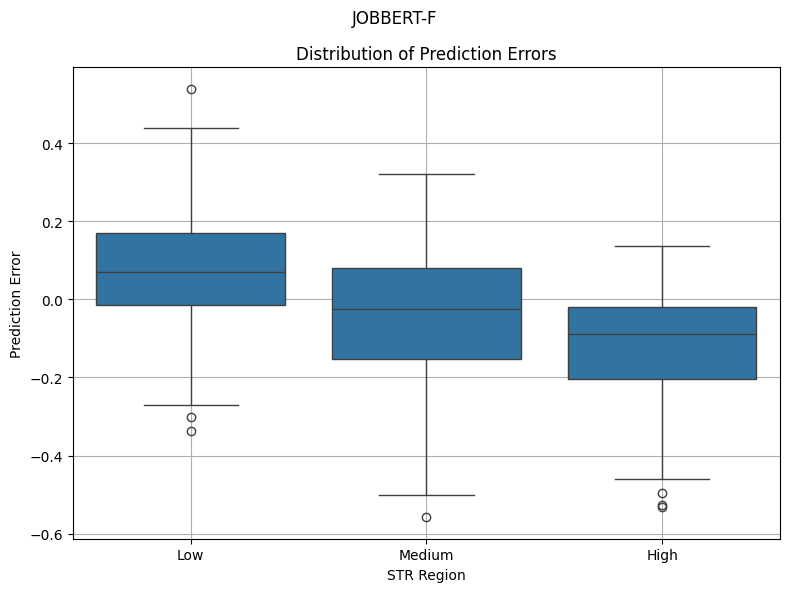}
\caption{Distribution of Prediction Errors for JOBBERT-F}
\label{fig:EVAL_ERROR_JOBBERT-F_BOXPLOT}
\end{figure}

JOBBERT-F exhibits fewer significant differences: low–medium and low–high contrasts are significant, but medium–high differences are not. This stability in medium and high STR regions may reflect the benefits of fine-tuning in reducing variance for more semantically similar pairs (Figure~\ref{fig:EVAL_ERROR_JOBBERT-F_BOXPLOT}).

\subsubsection{MPNET}
\begin{figure}[!htbp]
\centering
\includegraphics[width=7cm]{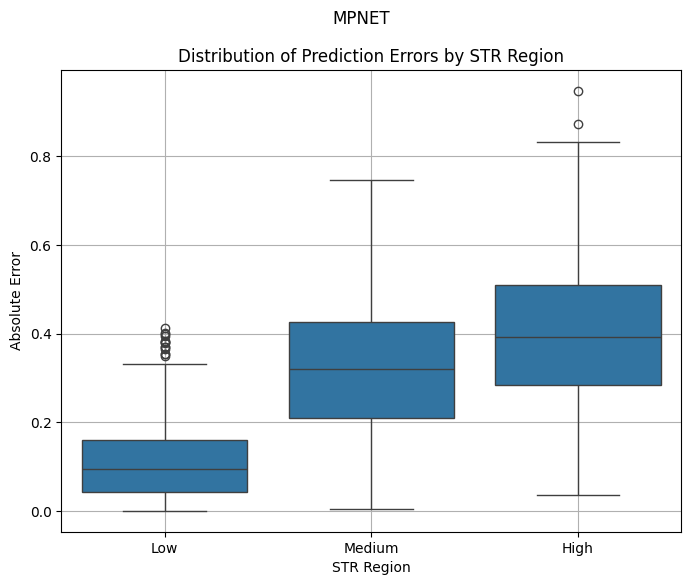}
\caption{Distribution of Prediction Errors for MPNET}
\label{fig:EVAL_ERROR_MPNET_BOXPLOT}
\end{figure}

MPNET shows strong, significant differences across all region pairs, with particularly large negative t-values for low–medium and low–high comparisons, indicating much lower errors in low STR compared to the other regions (Figure~\ref{fig:EVAL_ERROR_MPNET_BOXPLOT}). 

\subsubsection{MPNET-F}
\begin{figure}[!htbp]
\centering
\includegraphics[width=7cm]{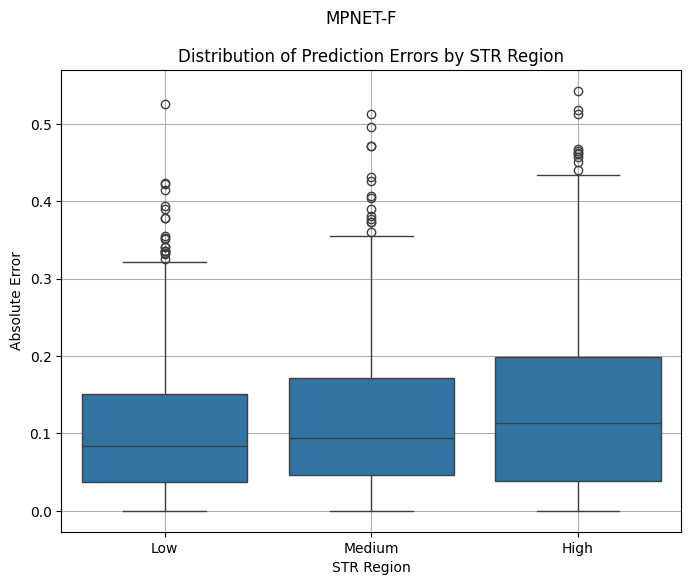}
\caption{Distribution of Prediction Errors for MPNET-F}
\label{fig:EVAL_ERROR_MPNET-F_BOXPLOT}
\end{figure}

MPNET-F shows strong, significant differences across all region pairs, with particularly large negative t-values for low–medium and low–high comparisons. However, it shows no significant difference between medium and high STR, suggesting more consistent performance at the higher end of the similarity spectrum (Figure~\ref{fig:EVAL_ERROR_MPNET-F_BOXPLOT}).

\subsubsection{MPNET-RGCN}
\begin{figure}[!htbp]
\centering
\includegraphics[width=7cm]{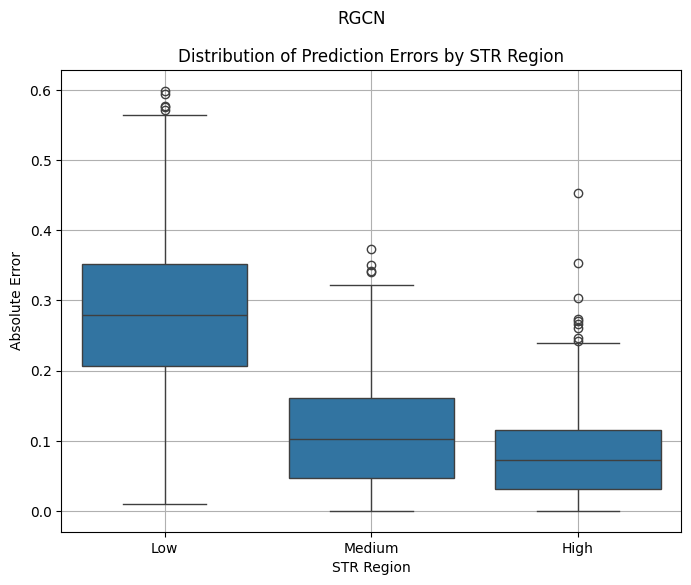}
\caption{Distribution of Prediction Errors for MPNET-RGCN}
\label{fig:EVAL_ERROR_RGCN_BOXPLOT}
\end{figure}

MPNET-RGCN demonstrates significant differences for all region pairs, with large positive t-values, implying higher errors in low STR and substantially lower errors in medium and high STR regions. This suggests that KG integration particularly benefits the model’s handling of semantically similar job pairs (Figure~\ref{fig:EVAL_ERROR_RGCN_BOXPLOT}).

Overall, these results confirm our hypothesis that models behave differently across STR regions, and that fine-tuning or KG integration can improve performance stability in specific ranges.

\subsection{Explainability Analysis}
A key motivation of our work is to provide \textbf{explainable} job-to-job matches through the integration of knowledge graphs, linking jobs and skills, and providing structured explanations.

\begin{figure}[!htbp]
\centering
\includegraphics[width=7cm]{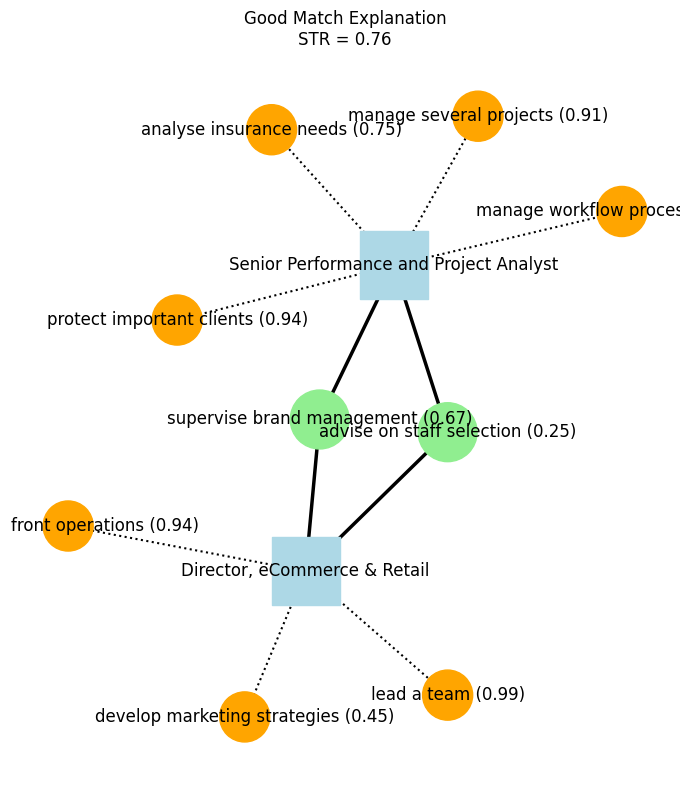}
\caption{Explanation Graph - Good Job Title match. The skill specificity (shown in brackets) is the inverse of the centrality degree}
\label{fig:EVAL_GOOD_MATCH}
\end{figure}

\begin{figure}[!htbp]
\centering
\includegraphics[width=7cm]{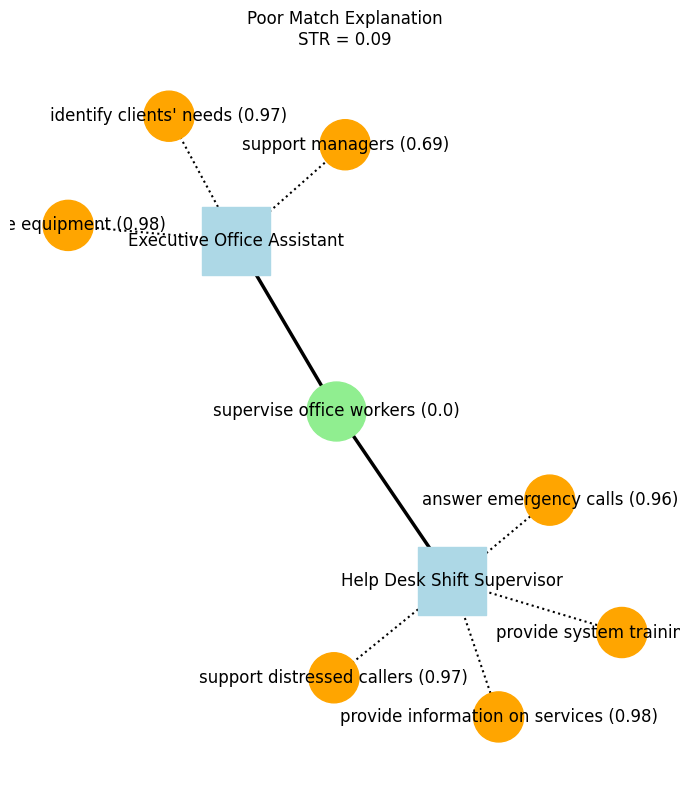}
\caption{Explanation Graph - Poor Job Title match. The skill specificity (shown in brackets) is the inverse of the centrality degree}
\label{fig:EVAL_POOR_MATCH}
\end{figure}

Figures ~\ref{fig:EVAL_GOOD_MATCH} and \ref{fig:EVAL_POOR_MATCH} present two illustrative cases: one good match and one poor match. For the high-STR pair \textit{``Senior Performance and Project Analyst''} vs. \textit{``Director, eCommerce \& Retail''}, the explanation highlights a shared set of highly-specific skills (e.g., \textit{``supervise brand management''} with specificity of 0.67). In contrast, the low-quality match \textit{``Executive Office Assistant''} vs. \textit{``Help Desk Shift Supervisor} is driven by overly generic skills (e.g., \textit{``supervise office workers''} with specificity of 0.0). Such explanations increase transparency by showing whether similarity arises from meaningful or spurious overlaps.

These results strengthen the claim that KGs improve explainability. Good matches can be justified by showing the specific shared skills that drive similarity, while poor matches can be diagnosed by revealing over-reliance on generic skills. 

\subsection{Practical Implications}
Our findings have important practical implications for downstream tasks such as re-ranking or candidate filtering. When irrelevant matches have already been discarded, the primary challenge lies in fine-grained differentiation among relevant alternatives. In such contexts, a model’s behavior in the medium to high STR range becomes especially critical (Figure~\ref{fig:EVAL_ERROR_RGCN_BOXPLOT}). Conversely, when filtering for dissimilar job pairs (e.g., in deduplication or anomaly detection tasks), performance in the low STR range is of greater interest (Figure~\ref{fig:EVAL_ERROR_MPNET_BOXPLOT}). This suggests that a general-purpose pre-trained Language Model can be used in the initial stages of the recommendation pipeline, with a transition to fine-tuned, domain-specific models at a later stage when more detailed distinctions are necessary.

Stratified evaluation also allows us to observe localized performance patterns and better characterize where models succeed or fail. This approach informs both model selection and training strategies — such as adapting loss functions to focus on under-performing regions or augmenting training data with examples from underrepresented STR bands.

Furthermore, by integrating Knowledge Graphs into the semantic matching process, we provide a structured reasoning path behind recommendations. This may improve transparency, build trust with stakeholders, and help recruiters justify why a specific job was recommended.

Although focused on job title and skill matching, our methodology can be applied to other domains such as academic paper recommendation, product matching, or legal case retrieval.

\subsection{Future Work}

Although our current approach provides a foundation for learning semantic representations of job titles using graph-based and textual signals, there are opportunities for future improvement and expansion.

First, our knowledge graph construction is limited to skill-based relationships. Incorporating additional semantic dimensions such as industry classifications, job seniority levels (e.g., “Lead Engineer” vs. “Intern”), and domain-specific contexts (e.g., “Data Scientist – Healthcare” vs. “Data Scientist – Finance”) would provide a more comprehensive representation of the job landscape. 

Second, the scope of our current model evaluation is constrained. We only explore a limited set of graph embedding models, focus exclusively on Cosine Similarity Loss, and implement a single negative sampling strategy. Future research should embrace a broader range of models, loss functions (e.g., contrastive loss, triplet loss), and negative sampling strategies to assess their effect on model performance.

Third, our work focuses exclusively on job-to-job matching. Extending the framework to job-to-resume and resume-to-job matching tasks could make it more useful in real-world recruitment systems.

Additionally, our approach currently relies on structured skill taxonomies such as ESCO \citep{EuropeanCommission2021ESCO} or O*NET \citep{onet}, which may limit generalizability in domains with less formalized ontologies. Also, we do not address multi-lingual job descriptions, which presents an important direction for future development, particularly for global labor markets.

Moreover, our reliance on weak supervision introduces potential label noise, especially in low-similarity cases, which raises concerns about general robustness and may limit the model's ability to generalize. 

Finally, we train and evaluate on a relatively small dataset, which may not capture the full variability of job title semantics across sectors or regions. Scaling the data set and introducing data augmentation techniques or semi-supervised learning methods could mitigate this limitation and improve the robustness of the model.


\section{Conclusion}

This study sets the direction for addressing a critical challenge in HR applications: the need for explainable and transparent recommendations. Although text embedding models like SBERT capture complex contextual semantics, they often lack interpretability. To overcome this limitation, we introduce a hybrid approach that combines Semantic Textual Relatedness (expressed by fine-tuned SBERT embeddings) with domain-specific knowledge graphs using Graph Neural Networks. This combination enables not only enhanced performance but also the ability to trace reasoning paths between matched job titles —- an essential feature for auditable and trustworthy decision-making in hiring contexts. As the use of Artificial Intelligence in recruitment systems expands, approaches that prioritize both semantic depth and interpretability will be key to ensuring fairness and user trust.

Ultimately, this research contributes to building intelligent, equitable and explainable recommendation systems that serve both candidates and employers in a dynamic and evolving labor market.


\begin{acknowledgments}
This research was partially supported by the Horizon Europe project GenDAI (Grant Agreement ID: 101182801) and by the ADAPT Research Centre at Munster Technological University. ADAPT is funded by Taighde Éireann – Research Ireland through the Research Centres Programme and co-funded under the European Regional Development Fund (ERDF) via Grant 13/RC/2106\_P2.

We would also like to thank the anonymous reviewers for their valuable feedback and constructive suggestions, which have helped to improve the quality and clarity of this work.

\end{acknowledgments}

\section*{Declaration on Generative AI}
In our literature review process, we employed Scite AI Research Assistant \citep{scite}, which allowed a more comprehensive review of the existing literature, ensuring that the sources included in this study were relevant and reliable. We utilized OpenAI Code Assistant \citep{openai_code_assistant} to accelerate code development and improve productivity during the implementation phase. The assistant was used to generate boilerplate code, troubleshoot and debug runtime errors, and explore alternative design patterns. We also used Grammarly \citep{grammarly} and Microsoft Copilot \citep{copilot2025} for paraphrasing and corrections of grammatical, syntactical, and other writing errors.
After using these tools/services, the authors reviewed and edited the content as needed and take full responsibility for the publication’s content. Generative AI tools were not used for data analysis, experimentation, or the formulation of hypotheses and conclusions.

\newpage
\bibliography{bibliography}

\section{Appendices}
\appendix
\section{Tables}\label{AppendixA}


\begin{table}[!htbp]
\centering
\begin{tabular}{|p{2cm}|p{3cm}|c|}
\hline
\textbf{Parameter}  & \textbf{Description} & \textbf{Value} \\
\hline

{High STR Range} &  Data region in which STR is considered 'high' & {[0.75, 1.0]}\\
\hline
{Medium STR Range} &  Data region in which STR is considered 'medium' & {[0.5, 0.75]}\\
\hline
{Low STR Range} &  Data region in which STR is considered 'low' & {[0, 0.5]}\\
\hline
{Number of Skills Per Job} &  Maximum number of skills assigned to a job & 10\\
\hline
{Job-Skill STR Threshold} &  Minimum STR value at which a skill is considered related to a job & 0.5\\
\hline
{Skill-Skill STR Threshold} &  Minimum STR value at which a child skill is considered related to a parent skill & 0.25\\
\hline
{Text Epochs} &  Number of epochs for SBERT fine-tuning & 5\\
\hline
{Graph Epochs} &  Number of epochs for KG model training & 15\\
\hline
\end{tabular}
\caption{Hyper-parameters of the proposed pipeline}
\label{table:EXP_HYPER_PARAMETERS}
\end{table}



\begin{table}[!htbp]
\centering
\begin{tabular}{|l|c|}
\hline
\textbf{Name}  & \textbf{Description} \\
\hline
{Environment} &  Google Colab\\
\hline
{Architecture} &  x86/x64\\
\hline
{CPU} &  8 CPUs @ 2.00GHz\\
\hline
{GPU} &  Tesla T4\\
\hline
{CUDA} &  Version 12.4\\
\hline
\end{tabular}
\caption{Runtime Environment Setup}
\label{table:EXP_SYSTEM_SPEC}
\end{table}

\begin{table}[!htbp]
\centering
\begin{tabular}{|p{2cm}|p{5cm}|}
\hline
\textbf{File Name}  & \textbf{File Description} \\
\hline
 Source Jobs & \textbf{Input}.  File 'source\_jobs.csv' is derived from a Kaggle dataset \citep{Koneru2024} and contains 14,000 records covering a wide range of professional roles: technical, creative, educational, financial, administrative, and operational.\\
\hline
 Source Skills &  \textbf{Input}. File 'source\_skills.csv' is derived from a list of 14,000 skills and competences \cite{EuropeanCommission2021ESCO} and a list of 50 skill categories \cite{Indeed2025.JobTitles}.\\
\hline
 Source Skill Hierarchy &  \textbf{Input}. File 'source\_skill\_hierarchy.csv' defines relationships between skill categories. For example, {\textquoteleft}Marketing{\textquoteright} and {\textquoteleft}Sales{\textquoteright} are grouped under {\textquoteleft}Sales \& Marketing{\textquoteright}\\
\hline
 Training Job Title Pairs &  \textbf{Output}. File 'train\_job\_title\_pairs.csv' is a training dataset. Each row includes anchor text (job title) which is the basis for comparison, sample text (job title), and STR score between the anchor and the sample\\
\hline
 Evaluation Job Title Pairs &  \textbf{Output}. File 'eval\_job\_title\_pairs.csv' is an evaluation dataset. Each row includes anchor text (job title) which is the basis for comparison, sample text (job title), and STR score between the anchor and the sample\\
\hline
\end{tabular}
\caption{Description of main data files}
\label{table:DATA_FILES}
\end{table}

\end{document}